\documentclass[conference, a4paper]{IEEEtran}
\IEEEoverridecommandlockouts

\usepackage{url}
\usepackage{upgreek}
\usepackage{amsmath,amssymb,amsfonts}
\usepackage{graphicx}
\usepackage{multicol}% http://ctan.org/pkg/multicols
\usepackage{arabtex}
\usepackage{utf8}
\usepackage{multirow}
\usepackage{array}
\usepackage{tabularx}
\setcode{utf8}

\usepackage{amsmath}
\usepackage{fancyhdr}
\newcommand{\correspondingauthor}{\thanks{{*}Corresponding author: Abdullah Can Algan}}

\begin{document}

\cfoot{\small{\text UBMK 2023 8th international conference on computer science and engineering (UBMK).}}
\title{A Use Case: Reformulating Query Rewriting as a Statistical Machine Translation Problem\\
%\thanks{XXX-X-XXXX-XXXX-X/XX/\$31.00~\copyright~2022 IEEE}
}

\author{\IEEEauthorblockN{1\textsuperscript{st} Abdullah Can Algan*\correspondingauthor}
\IEEEauthorblockA{
\textit{Huawei T\"{u}rkiye} \\
\textit{R\&D Center}\\
Istanbul, T\"{u}rkiye \\
0000-0002-8493-0883}
\and
\IEEEauthorblockN{2\textsuperscript{nd} Emre Y\"{u}rekli}
\IEEEauthorblockA{\textit{Huawei T\"{u}rkiye} \\
\textit{R\&D Center}\\
Istanbul, T\"{u}rkiye \\
emre.yurekli3@huawei.com}
\and
\IEEEauthorblockN{3\textsuperscript{rd} Aykut \c{C}ay{\i}r}
\IEEEauthorblockA{\textit{Huawei T\"{u}rkiye} \\
\textit{R\&D Center}\\
Istanbul, T\"{u}rkiye \\
0000-0001-9564-0331}

}

\maketitle

\begin{abstract}
One of the most important challenges for modern search engines is to retrieve relevant web content based on user queries. In order to achieve this challenge, search engines have a module to rewrite user queries. That is why modern web search engines utilize some statistical and neural models used in the natural language processing domain. Statistical machine translation is a well-known NLP method among them. The paper proposes a query rewriting pipeline based on a monolingual machine translation model that learns to rewrite Arabic user search queries. This paper also describes preprocessing steps to create a mapping between user queries and web page titles.          
\end{abstract}

\begin{IEEEkeywords}
search engine, query rewriting, statistical machine translation
\end{IEEEkeywords}

\begin{figure*}

\centering
\includegraphics[width=\textwidth]{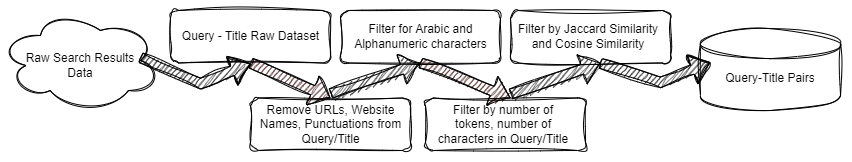}
\label{fig:mesh1}
\caption{Preprocessing diagram}
\centering
\end{figure*}

\section{Introduction}
Modern search engines are the most essential data sources created by world-wide users. From this perspective, a modern search engine is also considered a huge database where users can search and find content in. That is why, roughly a search engine has two important tasks such as indexing and retrieving the data or web page content. Data indexing is out of the scope of this paper.

Search engines focus on increasing the quality of retrieving relevant content as accurately and fast as possible. To do that, most of the well-known search engines leverage the improvements of the natural language processing models. The latest development in the machine and deep learning approaches provides users to search and find relevant web page contents in milliseconds. To improve the search experiences of the users, there are several important components such as query correction, rewriting and named entity recognition. This paper describes a use case that reformulates query rewriting as a statistical machine translation model.

Our main contributions in the paper can be summarized as follows:
\begin{itemize}
    \item To our best knowledge, we introduce the first statistical machine translation model to rewrite Arabic user queries using query-title pairs crawled from web search results.
    \item We present an Arabic language specific data pre-processing pipeline that struggles with the problems of queries due to being free text inputs.
\end{itemize}
The paper is organized as four main sections. Section \ref{sec:related-work} presents a literature review about query rewriting methods. Section \ref{sec:methodology} describes how we reformulate query rewriting as a machine translation problem. We briefly discuss the advantages of the proposed approach and its limitations in section \ref{sec:discussion}. Section \ref{sec:conclusion} concludes the paper.

\section{Related Work}
\label{sec:related-work}
Query rewriting is one of the most common techniques to improve search results in databases. Calvenese et al. define query rewriting for database management systems \cite{calvanese2000query}. Similar to database management systems, search engines need to rewrite user queries to retrieve more content aware results as fast as possible. In order to do that, search engines use some learning models that can find a semantic mapping between web documents and user queries \cite{he2016learning}. Grbovic et al. propose a content and context aware query rewriting module for search engines based on a novel query embedding algorithm \cite{grbovic2015context}.

Wu et al. introduce an interesting example for query rewriting model based on reinforcement learning \cite{wu2021conqrr}. They have compared their reinforcement model with a T5QR model which is based on a seq2seq transformer architecture. Text generation is a common task in NLP domain. With a breakthrough in transformer architecture, query rewriting is also a proper candidate for seq2seq models based on pre-trained transformers. Lin et al. \cite{lin2021contextualized} present a generative query rewriting model by using pre-trained T5 architecture. Another method is the few-shot generative approach for query rewriting. Yu et al. \cite{yu2020few} develop two methods based on rules and self-supervised learning, to generate weak supervision data and to finetune GPT-2 model with it. GPT-2 is a transformer-based large language model, which can perform zero-shot inference. Large language models also can handle context-independent queries effectively. 

However, seq2seq models and large language models have latency issues while performing inference on a search engine. Latency is one of the main challenges while working on search engines. Statistical models have the advantage over neural models considering latency. 

Mandal et al. \cite{mandal2019query} use synonyms for query rewriting. They take user queries and generate synonym candidates for the words in the query. They use a classifier to choose a synonym of the word from the candidates. Riezler and Liu \cite{riezler2010query} have used user query and document space to train an SMT model. Jun-Wei et al. \cite{bao2013query} have used user query and query-related document keywords to train an SMT model. Riezler and Liu \cite{riezler2010query} and Jun-wei \cite{bao2013query} have used similar methods for generating rewritten queries with different word alignment technics. In this work, the SMT model has been used with user queries and document titles, which is one of the different word alignment approaches.

\section{Methodology}
\label{sec:methodology}

\subsection{Problem Reformulation}
\label{subsec:reformulation}
Generally speaking, a machine translation contains two languages such as source and target languages. Let $M$ denote the machine translation model, $S$ denote the source language and $T$ denote the target language in a classical machine translation application. A machine translation model learns a mapping between sentence $s_{i}$ in $S$ and sentence $t_{i}$ in $T$ as shown in equation (\ref{eqn:mtm}). In this setting, $S$ and $T$ are different natural languages.

\begin{equation}
\label{eqn:mtm}
\begin{split}
    s \in S \\
    t \in T \\
    M(s) \rightarrow t
\end{split}
\end{equation}

\begin{equation}
\label{eqn:query_title}
\begin{split}
    q \in Q \\
    \tau \in \uptau \\
    M(q) \rightarrow \tau
\end{split}
\end{equation}

In order to develop a rewriting model based on machine translation settings, equation (\ref{eqn:mtm}) can be reformulated as a monolingual setting for source and target sentences, similarly to \cite{riezler2010query,yin2016ranking}. In setting (\ref{eqn:query_title}), $Q$ and $\uptau$ belong to the same language. In this way, model $M$ learns how to rewrite a sentence or query $q$ as sentence $\tau$.

\subsection{Statistical Machine Translation (SMT)}
Statistical Machine Translation is treating a translation task as a machine learning problem. The process relies on learning automatic translation using a parallel corpus. SMT is one of the key applications in the field of NLP. Core ideas of automatic machine translation go back decades ago. It has been studied widely and it interacts with many different areas such as statistics, linguistics, computer science, and optimization techniques. We will specify our system architecture and design choices but we will discuss them on a high level. Nevertheless, a good historical overview of the SMT approach is given in many survey papers \cite{hutchins2007machine,lopez2008statistical}.

As described in problem reformulation (Section \ref{subsec:reformulation}), SMT models are used to translate from a source language to a target language. Traditionally, the source and the target languages are different. However, in our system, we develop a monolingual SMT model. Considering our goal is being a query rewriting ability, we do not need two different languages. Therefore, both the source and target corpus are in Arabic.

We employ an open-source toolkit called Moses \cite{koehn2007moses} for Statistical Machine Translation. As the authors claimed, it is a complete toolkit for handling various translation tasks. From preprocessing the corpus to training translation systems, the tools included in Moses are capable enough. Important steps for SMT training are described as follows:

\textbf{Data preparation:} This step has two main objectives. One of them is to clean the parallel corpus by removing duplicate spaces and empty lines. Too short or too long sentences are also deleted. Since parallel corpus contains two different files, removing a sentence requires removing the corresponding line as well. The other objective is to lower the data sparsity which is done by lowercasing and tokenization.

\textbf{Word Alignment:} MGIZA \cite{gao2008parallel} is used for the word alignment between the source and the target languages. MGIZA is the multi-threaded implementation of GIZA++ \cite{och2003systematic}.

\textbf{Language modeling:} Language models are used for predicting the probability of a word based on the previous context. We trained a trigram language model and the standard probability formula of a sentence is shown in equation \ref{eqn:kenlmformula}. 
\begin{equation}
\label{eqn:kenlmformula}
    P(W) = P(w_1, w_2, w_3...w_N) \approx \prod_{i=1}^{N} p(w_i| w_{i-2}w_{i-1} )
\end{equation}

Language model implementation is based on KenLM \cite{heafield2011kenlm}.

\textbf{Tuning:} After completing the training phase, the model is tuned using Minimum Error Rate Tuning (MERT) \cite{och-2003-minimum} method.  100K query-title pairs are reserved for the tuning phase.

\textbf{Pruning the phrase table:} Our translation model is intended to be used in the real-time search pipeline. Therefore, our biggest constraint is latency. We have pruned the phrase table to lower the inference time. The phrase table contains Arabic phrase pairs and their co-occurrence frequencies. We have removed phrase pairs whose frequency is less than or equal to 3. In addition to performance gain, pruning is beneficial to increase phrase table quality because phrase tables have a considerable amount of noise due to alignment errors and noisy training data. Therefore, pruning allows us to eliminate these noisy phrase pairs.

\textbf{Inference:} After the training, the trained SMT model is tested on unseen data. The model tends to produce the same output as the input for the frequent phrases. We have implemented another step to increase the output variation. For this purpose, we have not relied on a single prediction from the model instead we gathered the top-5 outputs for each test sentence. Then, we used these outputs iteratively. Starting from the first output, we have tested if a prediction is exactly the same as the input. If so, we check the next prediction. In the end, if all the outputs are the same, we finally used the fifth-best prediction.

\subsection{Dataset}

From search engines, $\sim$28M Query – Title pairs have been collected for the Arabic language.
%Since these queries and titles are private information, there have been shared only some examples in order to explain the preprocessing steps.
After preprocessing steps have been applied to this raw data, remained 6,350,718 query-title pairs.

\textbf{Preprocessing steps:}
The thresholds that have been used for the following steps have been determined by native Arabic annotators by analyzing random query-title pairs' sub-sample sets.

Punctuation, URL and website names have been replaced with white space in queries or titles if they exist. These parameters do not have any contribution to this work. Titles, which come from search results, may include website names. User queries and re-written queries have an effect on the search results. To prevent boosting any website and to protect the user’s intention, website names  also have been replaced with white space in the titles.

% \begin{figure}[h]
% \caption{Preprocessing diagram}
% \centering
% \includegraphics[width=0.9\linewidth]{preprocessing_diagram.png}
% \label{fig:mesh1}
% \centering
% \end{figure}

Examples of the titles including website name: 

\begin{itemize}
\item \<الأخبار العاجله والمؤكده - الصفحة الرئيسية فيسبوك>
\item[-] En: "Urgent and confirmed news - Facebook homepage"
\item \<المكتب  مسلسل أمريكي  - ويكيبيديا>
\item[-] En: "The Office American Series - Wikipedia"
\end{itemize}

Examples of the queries that are mixed queries:
\begin{itemize}
\item \<اغنية حبيتك يابنت الناس mp3>
\item[-] En: "Song of Your Love, Daughter of People mp3"
\item \<تسريحات الشعر simple للأطفال>
\item[-] En: Simple hairstyles for children
\end{itemize}
In addition to them, raw data contains all the search results for a query, however, in order to keep the semantic relation between query and title pairs higher, only the top 5 search results for a query have been considered.

If the number of characters in queries or titles is less than 20 or the number of tokens in queries or titles is less than 3, then query-title pairs have been removed. When these short queries or titles have been considered, user intention can be lost. If queries or titles include repeated tokens more than 3 times, these tokens have been replaced with one white space.

The main assumption of this work is that queries and titles have a relation semantically and they can be used instead of each other. To catch semantically similar phrases, the number of tokens in queries and titles must be close to each other. Therefore, if the difference between the number of tokens in the query and in the title is bigger than 3, the query-title pair has been eliminated. 

Queries and titles may include words in other languages instead of Arabic. They also may include some special characters which are like "\fontencoding{T1}\selectfont ß", "\fontencoding{T1}\selectfont ä", "\<بِسْــــــــــــــــــــــمِ>" etc. To cover both of these cases, firstly, the alphanumeric characters filter, then, the Arabic characters filter have been applied. The thresholds are 0.9 and 0.7 for the alphanumeric characters filter and the Arabic characters filter, respectively.

%the threshold has been determined as 0.9, and for the Arabic characters filter, the threshold has been determined as 0.7. 

After these cleaning processes, Jaccard index and cosine similarity score have been calculated for semantic analysis of the query-title relation. 

For Jaccard index, the NLTK-SnowballStemmer function has been used. Stems for each token have been extracted from queries and titles. In equation (\ref{eqn:jaccard}), Q represents the stem list of the words in the query while T represents the stem list of the words in the title. 

\begin{equation}
\label{eqn:jaccard}
    J(Q, T) = \frac{| Q \cap T |}{|Q \cup T| }
\end{equation}

%\[J(Q, T) = \ \frac{| Q \cap T |}{|Q \cup T| }\]

For each query-title pair, Jaccard index has been calculated and if it is less than 0.35, the pairs have been eliminated. 

For the cosine similarity index, RoBERTa based, pre-trained sentence transformer model's \footnote{https://huggingface.co/symanto/sn-xlm-roberta-base-snli-mnli-anli-xnli} embeddings have been used. The lower index is 0.5, while the upper index is 0.9. In equation (\ref{eqn:cosine}), Q represents query embeddings while T represents title embeddings.
\begin{equation}
\label{eqn:cosine}
Cos\theta = \frac{| \vec{Q} \times \vec{T} | } {||\vec{Q}|| ||\vec{T}||}  
\end{equation}
% \[Cos\theta = \frac{| \vec{Q} \times \vec{T} | } {||\vec{Q}|| ||\vec{T}||} \]

Both the Jaccard index and cosine similarity score have been calculated to measure the distance between the query and title. If query and title are the same or they have less relation with each other based on these scores, then, the pair have been eliminated from the data. 

\section{Discussion}
\label{sec:discussion}

%IN-PROGRESS.
In an effort to test whether our rewriting corrupts the local phrases, we have tested the model's prediction on localized queries. Localized queries are test sets that include local entities, dialect words, and transliterations. Arabic is spoken across many different regions. Therefore, it is another challenge to preserve region-specific phrases while rewriting the query. Each Arabic-speaking region has different language usage styles. Since our model is not context-aware, sometimes it outputs a query that has localization issues.

We have observed an increase in the end results. Nevertheless, our system design has some limitations. A purely statistical approach depends on the training corpus meaning that larger texts might enable better results. However, it is inevitable that the data in the production system will differ from the training data at one point. There are several reasons for this possible difference. Firstly, the language itself is constantly evolving. New terms are introduced in our daily life. Also, foreign words are being used for new concepts. In a nutshell, the ratio of out-of-vocabulary words should be monitored and the system updates must be implemented accordingly.

%\subsection{Model outputs:}

\begin{table}[]
    \centering
    \caption{Examples that are annotated as bad from the model outputs}
    \begin{tabular}{ |c|p{0.6\linewidth}|p{0.2\linewidth}| }
    \hline
        & Examples & Error Types \\
    \hline
         \multirow{4}{1em}{1} & User query: \<كلمه مساء عن السلام> & \multirow{4}{4em}{Change intention or meaning} \\
         & En: Evening speech on peace & \\
         & Rewritten query: \<كلمات عن المساء> & \\
         & En: Words about the evening & \\
    \hline
         \multirow{5}{1em}{2} & User query: \<العربات الصغيرة الموسم 2 الحلقة 11> & \multirow{5}{4em}{Change numbers} \\
         & En: Buggies season 2 episode 11 &  \\
         & Rewritten query: & \\
         & \<العربات الصغيرة الموسم 2 الحلقة 9> &  \\
         & En: Buggies season 2 episode 9 & \\
    \hline
         \multirow{4}{1em}{3} & User query: \<الطقس في الرياض طقس العرب> & \multirow{4}{4em}{Delete words} \\
         & En: Weather in Riyadh Arabia weather & \\
         & Rewritten query: \<الطقس ساعة بساعة> & \\
         & En: Weather hour by hour & \\
    \hline
         \multirow{5}{1em}{4} & User query: \<مشهور وناسه للحلويات جازان> & \multirow{5}{4em}{Change location} \\
         & En: Famous and its people for sweets Jazan & \\
         & Rewritten query: & \\
         & \<مشهور وناسه للحلويات فرع صبيا> & \\
         & En: Famous and its people for sweets Sabya branch & \\
    \hline
        \multirow{5}{1em}{5} & User query: \<دائرة الاحصاء والتنمية المجتمعية> & \multirow{5}{4em}{Add redundant words or adjectives} \\
         & En: Department of Statistics and Community Development & \\
         & Rewritten query: & \\ 
         & \<دائرة الإحصاء والتنمية المجتمعية   الاقتصادي> & \\
         & En: Department of Statistics and Community Economic Development & \\
    \hline
         \multirow{4}{1em}{6} & User query: \<مسلسل جمان نور الغندور حلقه 1> & \multirow{4}{4em}{Normalization problem} \\
         & En: Juman Nour Al-Ghandour series, episode 1 & \\
         & Rewritten query: \<مسلسل جمان الحلقة 1 الاولي> & \\
         & En: Juman series, episode 1, the first & \\
    \hline
    
    \end{tabular}
    \label{tab:bad_examples}
\end{table}

At the end of this work, native Arabic annotators have evaluated the SMT model outputs for the test set as "Good" and "Bad". The test set has been generated from the search queries randomly. Good examples can be seen in Table \ref{tab:good_examples} and bad examples can be seen in Table \ref{tab:bad_examples} with their error types.

The main annotation criterion is to protect user intention. If the rewritten query has changed the user intention or the meaning of the user query, then, it has been annotated as bad. In addition to this criterion, if the rewritten query has been generated grammatically wrong, then it also has been annotated as bad. 

In the good cases, it can be seen that in the rewritten queries, user intention has been protected and also they have been generated grammatically correct. Furthermore, when they have been analyzed, the SMT model is disposed to rewrite queries as more general like by deleting the years in the Table \ref{tab:good_examples} (Example 1), more specific like by adding words in the Table \ref{tab:good_examples} (Examples 3, 4 and 5) and by changing location in the Table \ref{tab:good_examples} (Examples 2 and 6). Moreover, the SMT model has rewritten queries by replacing words in the queries with their synonyms or alternatives. 

For the bad cases, the annotation team has analyzed the error types which have been categorized into 6 groups. Changing query intention or meaning, the numbers, and the location and deleting words are actually sub-groups of the first annotation criterion. The other error groups, adding redundant words or adjectives, and normalization problems, are the sub-groups of the second annotation criterion. 

\begin{table}[]
    \centering
    \caption{Examples that are annotated as good from the model outputs}
    \begin{tabular}{ |c|p{0.8\linewidth}| }
    \hline
        & Examples \\
    \hline
         \multirow{4}{1em}{1} & User query: \<الدوري السعودي 2021 المحترفين> \\
         & En: Saudi League 2021 Professionals  \\
         & Rewritten query: \<دوري المحترفين السعودي>  \\
         & En: Saudi Professional League \\
    \hline
         \multirow{4}{1em}{2} & User query: \<دخول الجامعة الاسلامية> \\
         & En: Entering the Islamic University  \\
         & Rewritten query: \<الجامعة الإسلامية بالمدينة المنورة>  \\
         & En: The Islamic University of Madinah \\
    \hline
         \multirow{4}{1em}{3} & User query: \<منصة مدرستي> \\
         & En: Madrasati platform \\
         & Rewritten query: \<رابط منصة مدرستي> \\
         & En: Madrasati platform's link \\
    \hline
         \multirow{4}{1em}{4} & User query: \<سعر الدولار> \\
         & En: The price of dollar \\
         & Rewritten query: \<سعر الدولار اليوم> \\
         & En: The price of dollar today \\
    \hline
         \multirow{4}{1em}{5} & User query: \<حلول خامس> \\
         & En: Halul-the fifth grade \\
         & Rewritten query: \<حلول خامس ابتدائي> \\
         & En: Halul-fifth elementary grade \\
    \hline
         \multirow{4}{1em}{6} & User query: \<ماكدونالدز في المملكة العربية السعودية> \\
         & En: McDonlad's in Suadi Arabic \\
         & Rewritten query: \<ماكدونالدز رياض> \\
         & En: McDonald's in Riyad \\
    \hline
    
    \end{tabular}
    \label{tab:good_examples}
\end{table}

\section{Conclusion and Future Work}

\label{sec:conclusion}
In this study, query rewriting is treated as a machine translation task. Traditional translation tasks are using bilingual data aiming a translation between different language pairs. However, we use a monolingual translation model to rewrite user-generated queries. There are many alternative methods to develop a translation system. We have chosen the Statistical Machine Translation approach. We have considered some aspects including the training data size, output quality, and inference speed. 

Low inference time is one of our most important requirements. Latency issues are handled with some optimization techniques. Firstly, trigram language models are converted to binary format. It speeds up the initial loading time significantly. Pruning the phrase table is very beneficial since it reduces the search space. Also, pruned phrase table requires much less memory for storage.

Although we have observed a significant performance increase, there remains some future work. As a data-driven method, SMT could take advantage of more training data. One of our plans is to gather larger monolingual parallel corpora. Also, our dataset is created automatically so there might be noisy query-title pairs. We are planning to eliminate those kinds of pairs with automatic evaluation metrics to increase the quality of the training data. These future works are directly related to our current approach meaning that statistical methods are limited to training data. Since we are trying to rewrite a query by preserving the intention, we need different variations. Encoder-decoder based Neural Machine Translation (NMT) methods might be useful because they make use of neural embeddings. 

\bibliographystyle{ieeetr}
\bibliography{reference}

\end{document}